\documentclass[openacc]{rstransa}

\usepackage{amsmath,amsfonts}
\usepackage{algorithmic}
\usepackage{algorithm}
\usepackage{array}
\usepackage{textcomp}
\usepackage{graphicx}
\usepackage{multirow}
\usepackage{tcolorbox} 
\usepackage[utf8]{inputenc} 
\usepackage[T1]{fontenc}    
\usepackage{url}            
\usepackage{booktabs}       
\usepackage{nicefrac}       
\usepackage{microtype}      
\usepackage[export]{adjustbox}
\usepackage{subfig}
\usepackage{lmodern} 
\usepackage{listings}

\definecolor{keywords}{RGB}{127,159,191}
\definecolor{comments}{RGB}{147,112,219}
\definecolor{red}{RGB}{220,50,47}
\definecolor{green}{RGB}{138,226,52}
\definecolor{dkgreen}{rgb}{0,0.6,0}
\definecolor{gray}{rgb}{0.5,0.5,0.5}
\definecolor{mauve}{rgb}{0.58,0,0.82}
\definecolor{mygray}{rgb}{0.9,0.9,0.9}
\definecolor{LightGray}{gray}{0.95}

\floatname{algorithm}{Procedure}

\DeclareMathOperator*{\argmin}{\textup{arg min }}

\usepackage{tabularx}

\titlehead{Research}

\history{
\textbf{Preprint / Author Accepted Manuscript (AAM)}.
This version has been accepted for publication in \textit{Philosophical Transactions A}. It has not undergone final copyediting, typesetting, or publisher formatting. The final version of record will be accessible from the Royal Society's website once available.
}

\begin{document}

\title{Towards symbolic regression for interpretable clinical decision scores}

\author{
Guilherme Seidyo Imai Aldeia$^{1}$, Joseph D. Romano$^{3}$, Fabricio Olivetti de Franca$^{1}$,  Daniel S. Herman$^{2}$, William G. La Cava$^{4}$}

\address{$^{1}$Federal University of ABC, SP, Brazil\\
$^{2}$Department of Pathology and Laboratory Medicine, University of Pennsylvania, PA, US\\
$^{3}$Institute for Biomedical Informatics, University of Pennsylvania, PA, US\\
$^{4}$Boston Children's Hospital, Harvard Medical School, MA, US}

\subject{xxxxx, xxxxx, xxxx}


\corres{William G. La Cava\\\email{William.LaCava@childrens.harvard.edu}}

\begin{abstract}
    Medical decision-making makes frequent use of algorithms that combine risk equations with rules, providing clear and standardized treatment pathways.   
    Symbolic regression (SR) traditionally limits its search space to continuous function forms and their parameters, making it difficult to model this decision-making. 
    However, due to its ability to derive data-driven, interpretable models, SR holds promise for developing data-driven clinical risk scores.
    To that end we introduce Brush, an SR algorithm that combines decision-tree-like splitting algorithms with non-linear constant optimization, allowing for seamless integration of rule-based logic into symbolic regression and classification models.
    Brush achieves Pareto-optimal performance on SRBench, and was applied to recapitulate two widely used clinical scoring systems, achieving high accuracy and interpretable models. 
    Compared to decision trees, random forests, and other SR methods, Brush achieves comparable or superior predictive performance while producing simpler models.
\end{abstract}


\begin{fmtext}
\end{fmtext}
\maketitle

\section{Introduction}~\label{sec:introduction}
Symbolic Regression (SR) is a supervised regression machine learning method that aims to discover a mathematical expression and its parameters from data.

Since SR was first proposed~\cite{koza1994genetic}, it was applied in various fields, including physics~\cite{udrescu_ai_2020, lalande2023a, doi:10.1126/sciadv.aay2631, Makke2024, Angelis2023}, medical informatics~\cite{la_cava_flexible_2023, Wilstrup2022, 10.1145/3205455.3205604}, aerospace engineering~\cite{LACAVA2016892, windturbinedynamics}, and material science~\cite{Wang_2019}.
The main appeal of SR is its potential to produce intrinsically interpretable solutions, often crucial for scientific discovery in basic sciences or when prediction models must justify their recommendations,~\textit{e.g.}, in health care contexts, \textit{c.f.} the United States~\cite{FDA} guidelines for clinical decision support software.

While promising for healthcare, purely mathematical expressions can be a drawback in practice due to the mathematical literacy they require, making them less attractive than decision trees~\cite{breiman1984classification}, which are often preferred when interpretability is needed~\cite{10.1214/21-SS133}.
In the healthcare context, widely used triage scoring systems in the U.S., such as CART~\cite{churpek2012derivation}, are based on physiological parameters and can be promptly evaluated to predict urgent deterioration of patients \cite{subbe2001validation}.
Scoring systems require interpretability, consisting of simple decision rules, and have been associated with better hospital outcomes~\cite{tan2022modified}.
They are common clinical decision tools because they provide objective, quantifiable measures for decision-making and have historically been developed manually \cite{guidetti2024symbolic}.
These systems could be explored with the aid of big data and artificial intelligence \cite{xie2022benchmarking}, especially through the usage of electronic health records (EHR). 

Contemporary SR algorithms that achieve good results in modeling regression equations for real-world and physics data~\cite{srbench} relies local parameter optimization~\cite{aldeia2025call}, but a limitation is not being able to incorporate split-wise operations, due to the difficulty of optimizing such parameters with existing optimization methods.
This task remains underexplored in SR.

\begin{figure}[htb]
    \centering
    \includegraphics[width=0.9\linewidth]{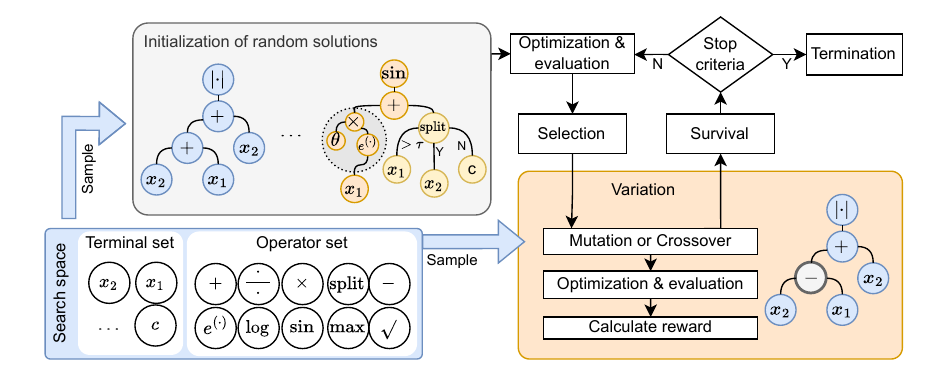}
    \caption{Brush Overview. Nodes are sampled from the search space to build mathematical expression trees, including those with split operations. A randomly generated population of expressions goes into an evolutionary loop.
    The set of nodes are used to create variations on the population. After several generations, the final solution is selected from the evolved individuals.}
    \label{fig:brush-overview}
\end{figure}

We propose a genetic programming symbolic regression algorithm named \textit{Brush}, aiming to bridge the gap between symbolic prediction models and split-wise operations, along with non-linear parameter optimization.
Brush addresses the multi-objective optimization problem of simultaneously maximizing performance and minimizing model complexity.
It learns split operations at any point within the expression while maintaining compatibility with non-linear optimization, even in the presence of discontinuities. Figure~\ref{fig:brush-overview} depicts our proposed algorithm.

Our experiments are two-fold.
In the first set of experiments, we validate our method and place it in the context of other SotA approaches. We ran Brush on $122$ real-world and $130$ synthetic physics problems from SRBench~\cite{srbench}, reporting $R^2$ and accuracy solution of governing physics equations. Brush achieved Pareto-optimal performance in SRBench and, compared to other SR methods, produced significantly smaller expressions. It achieved $R^2 > 0.999$ in more than half of the runs, even under high amount of noise.

In the second set of experiments, we applied Brush to regression and classification tasks using EHR data from Beth Israel Deaconness Medical Center. We extracted data for $10,000$ individuals from the MIMIC-IV~\cite{johnson2023mimic} emergency department table, including vital signs and demographics collected at triage, and applied Brush to the regression task of learning triage scores --- namely CART, and MEWS.
Brush produced competitive results with much smaller expressions. We then modify the regression task to a classification scenario, aiming to identify patients at high risk of catastrophic deterioration based on vital signs.
Brush successfully generated small models with simple split rules to flag high-risk patients.

Our results show that Brush is competitive with SotA symbolic regression algorithms while incorporating split-wise equations, and can also support decision-making by generating models that reflect the features and logic of ground-truth systems.
The algorithm is promising both as a first-principles modeling tool and as a method for learning interpretable classification models, expanding the practical applications of symbolic regression.

\section{Related work}~\label{sec:related-work}

The idea of mixing decision trees with linear regression models as leaves has been previously explored~\cite{Gama2004, rusch2013gaining}, but recent methods have integrated SR with decision trees~\cite{ZHANG2022101061, piecewisesr, fong2024symbolic}.
PS-Tree~\cite{ZHANG2022101061} builds decision tree-like structures with symbolic regression models as leaves and achieves competitive $R^2$ performance, but at the cost of overly large models. It lacks parameter optimization, relying on randomly generated constants.
Another method similar to PS-Tree generates decision trees with symbolic models as leaves~\cite{piecewisesr}, performing well on problems with or without required splits.
A different approach, SREDT~\cite{fong2024symbolic}, uses splits derived from an SR algorithm. SREDT outperforms traditional decision trees but also lacks parameter optimization.

Optimizing free parameters remains challenging, as random constants hinder convergence and fail to reach global optima~\cite{WANG20042453}, with evolutionary methods alone converging slowly~\cite{8393325}. This burden can be reduced using optimization methods. Recent SR algorithms apply Levenberg-Marquardt (LM)~\cite{levenberg1944method, marquardt1963algorithm} to tune model parameters~\cite{ITWithCoeffsOptimization, ParameterIdentificationForSR, PrioritizedGrammarEnumeration, operon}, which has proven effective.
Some pre-trained transformer-based methods also struggle with constants, highlighting the need for nonlinear optimization to improve model accuracy~\cite{kamienny_end--end_nodate, UnifiedFrameworkForDeepSR, shojaee_transformer-based_nodate}.

Previous work by La Cava \textit{et al.}~\cite{la_cava_flexible_2023} modified the FEAT~\cite{la2018learning} SR algorithm to perform logical comparisons, and it was applied to learn linear combination of meta-features in a logit function for developing computable phenotypes for hypertension.
Other SR applications in healthcare include using SR for feature engineering inputs to classifiers for hearth failure prediction~\cite{Wilstrup2022}, or learn meta-features from pediatric patients to predict scores from CT scans~\cite{10.1145/3205455.3205604}.

\section{Brush: mixing split-wise functions and parameter optimization}~\label{sec:brush}

Brush is a symbolic regression algorithm capable of learning expressions with split-wise operations, without being constrained to a decision tree-like structure.
Its splits are compatible with non-linear parameter optimization methods, which are lacking in current algorithms.
Additionally, Brush integrates several modern strategies within the genetic programming framework to efficiently explore the search space.
This section introduces the split node and its optimization approach, followed by a description of the mechanisms integrated into Brush.

Let a dataset be a set of $d$ observed points $\left\{(\mathbf{x}_i, y_i) \right\}_{i=1}^{d}$, where $\mathbf{x}$ is a $n$-dimensional feature vector, and $y$ is the target value. 
We denote the feature matrix as $\mathcal{X} = \left [ \mathbf{x}_{1}, \mathbf{x}_{2}, \ldots, \mathbf{x}_{d} \right ]' \in \mathbb{R}^{n \times d}$, and the vector of target values as $\mathbf{y} = \left [ y_{1}, y_{2}, \ldots, y_{d} \right ] \in \mathbb{R}^d$. 
Brush searches for $\hat{f}(\mathcal{X}, \hat{\theta}) : \mathbb{R}^n \rightarrow \mathbb{R}$ such that $\hat{f}(\mathcal{X}) \approx \mathbf{y}$, where $\hat{f}$ is a mathematical expression represented as a tree.

Brush introduces a specialized \emph{split node operator}, inspired by classical decision trees \cite{breiman1984classification} --- a split is a predicate determining whether to evaluate the left or right branch based on a boolean output.
During learning, the algorithm exhaustively tests combinations of feature $x_i$ and threshold $\tau$, selecting the configuration $[x_{i^*} > \tau^*]$ that minimizes the total variance of the target values in the resulting partitions, $\mathbf{y}_{[x_{i^*} > \tau^*]}$ and $\mathbf{y}_{[x_{i^*} \leq \tau^*]}$.

Our split node has three branches: one real-valued subtree $f_c$ as the condition input, and one subtree each for the true and false cases $f_T$ and $f_F$.
Each branch is another Brush tree.
The conditional compares $f_c$ with a learnable weight $\tau$ to determine the data flow between $f_T$ and $f_F$.
The optimal $\tau^*$ minimizes the sum of target variances on either side of the split, effectively performing one-dimensional clustering:

\begin{align}~\label{eq:split_threshold}
    \tau^* = & \underset{\tau}{\text{min}} \left ( \frac{\text{Var}(\mathbf{l})}{|\mathbf{l}|} + \frac{\text{Var}(\mathbf{r})}{|\mathbf{r}|} \right ) \\ 
     & \text{such that } \text{min}(f_c(\mathcal{X}, \mathbf{\theta})) < \tau < \text{max}(f_c(\mathcal{X}, \mathbf{\theta})) \\
     & \text{where } \mathbf{l} = [y_i : \mathbf{y} | f_c(\mathbf{x}_i, \mathbf{\theta}) \leq \tau], \\
     & \hspace{2.7em} \mathbf{r} = [y_i : \mathbf{y} | f_c(\mathbf{x}_i, \mathbf{\theta}) > \tau].
\end{align}

This operator masks the dataset, so downstream operations are applied only to a subset of the training data. Restricting subtrees to a subset of the data allows the resulting subtree to be more precise and enables faster parameter optimization.
In Brush, the initial population can optionally consist solely of split nodes, which are then progressively replaced by mathematical expressions through genetic programming whenever predictive performance is improved.
Each node is assigned an innate weight, which can be toggled during the search. In the initial population, only terminal weights are toggled on.
This extends Operon~\cite{operon} mechanism that assigns weights exclusively to terminals.
Let the node weights and any constants in an expression be adjustable parameters of the function ($\theta$).
Define the residual error as the result of the function $\hat{f}$ with parameters $\mathbf{\theta} \in \mathbb{R}^p$ as $\textup{H}(\mathbf{\theta}) = \widehat{f}(\mathcal{X}, \mathbf{\theta}) - \mathbf{y}$.
Then, the optimization problem is done by an iterative process of gradient descent to minimize the mean squared error described by Levenberg-Marquardt~\cite{levenberg1944method,marquardt1963algorithm}:
 
\begin{equation}~\label{eq:lm}
    \mathbf{\theta}^* = \underset{\mathbf{\theta}}{\argmin} \frac{1}{2}||H(\mathbf{\theta})||^2.
\end{equation}

Since any subtree is a valid Brush tree, we can isolate optimization to subtrees.
We propose the following heuristic for optimization with split nodes.
First, we optimize the conditional subtree $f_c$, then find the best threshold $\tau^*$ that clusters the data into two groups with minimal variance.
Finally, we fit the remaining expression, ignoring the already optimized parameters and passing only the data subset matching the condition to each corresponding branch.
We propose a \textit{greedy split node}, where the conditional is a single feature and the threshold is learned as in decision trees, and a \textit{flexible split node}, where the evolutionary framework generates a sub-expression and only its threshold is optimized.

\begin{figure}[htb]
    \centering
    \includegraphics[width=0.95\linewidth]{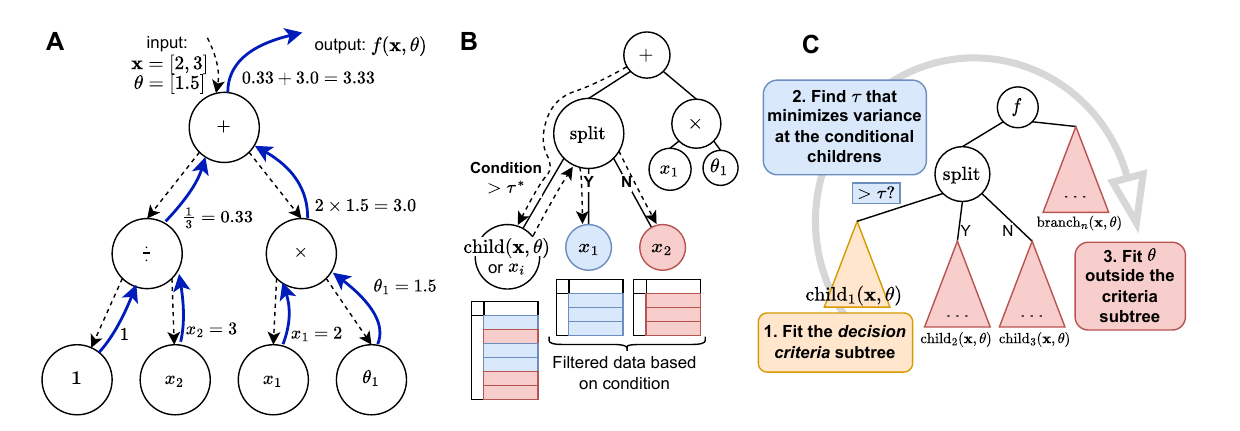}
    \caption{Brush evaluation, split, and optimization.
    (\textbf{A}) Evaluation starts at the root, recursively propagating input $\mathbf{x}$ and parameters $\theta$, applying each node's symbol to its children.
    (\textbf{B}) Splits can occur anywhere, with the conditional subtree directing data flow based on the condition.
    (\textbf{C}) Optimization with splits in three steps: optimize $\text{child}_1$; find the $\tau$ minimizing $y$ variance across $\text{child}_2$ and $\text{child}_3$; then fit the rest of the tree, ignoring already fitted parameters.}
    \label{fig:brush-evaluation}
    \vspace{-1.0em}
\end{figure}

Figure~\ref{fig:brush-evaluation} illustrates (A) tree evaluation, (B) how a split node partitions data across subtrees, and (C) the overall evaluation and optimization process. The split nodes minimize the sum of target variances between the two subtrees. The optimization is done in three steps: first, we optimize the decision criterion (Eq.~\ref{eq:lm}) --- which may be a single feature with a threshold or a subtree --- then determine the optimal threshold for the split (Eq.~\ref{eq:split_threshold}), and finally fit the remaining parameters (Eq.~\ref{eq:lm}) while keeping the split condition and threshold fixed.

Brush uses genetic programming (GP) to optimize overall fitness of a population of candidate models.
The fitness is defined as a vector of objectives we want to optimize. Specifically, we use the Mean Squared Error $\text{MSE}(\mathbf{\widehat{y}}, \mathbf{y}) = \frac{1}{d} \sum_{i=1}^{d} \left ( \widehat{y}_i - y_i \right )^2$ and the linear complexity to concurrently prioritize simplicity.
For a node $n$ with $k$ arguments, the linear complexity is the sum of its children's linear complexities with the complexity of the node itself $c_n$, that is $C(n) = c_n + (\sum_{a=1}^k C(a))$, discouraging the use of operators with high complexity values.

The GP loop starts with a randomly initialized population, then iteratively performs selection, variation, and survival steps.
\textit{Selection} chooses candidates based on fitness to undergo \textit{variation}, generating offspring that inherit characteristics from parents.
Variation doubles the population size, which returns to its original size during \textit{survival}, where selection pressure favors better solutions, with probabilities proportional to their fitness.
Brush uses $\epsilon$-lexicase~\cite{la_cava_epsilon-lexicase_2016} for selection, and non-dominated sorting~\cite{poliFieldGuideGenetic2008,NSGA-II} (NSGA-II) for survival, which are algorithms proven to yield better convergence performance than other selection and survival mechanisms.

Finally, we apply inexact simplification~\cite{inexact_simplification} as a post-processing step. This technique identifies and replaces large sub-trees with simpler, approximately equivalent alternatives by comparing prediction vectors. This allows us to compress models while preserving their performance.

We implemented $6$ different mutation operators and a subtree crossover. 
\textit{Toggle weight on/off} randomly enables or disables a learnable weight on a node.
\textit{Subtree} replaces one node with a random sub-tree, generated with PTC2. \textit{Point} replaces one random node with a new one of same arity. \textit{Delete} removes one node, keeping its children. \textit{Insert} creates a new node. The \textit{crossover} randomly switches two sub-trees between the selected parents. 

Finally, we extend Brush to classification tasks by introducing a dedicated classifier mode. In this setting, a logistic regression node is fixed at the root of the expression tree. We also enforce the presence of an offset parameter, which can be optimized using the same non-linear optimization methods, without requiring any modifications to the algorithm, and broadens the applicability of Brush to practical problems. 

\section{Performance on the Symbolic Regression Benchmark}~\label{sec:methodology}

In our first set of experiments, we demonstrate that Brush can achieve competitive performance on symbolic regression benchmarks. The SR community has adopted SRBench~\cite{srbench} as the standard benchmark for symbolic regression. It is composed of two tracks: \textit{black-box} and \textit{ground-truth} problems.
The former are $122$ datasets derived from the Penn Machine Learning Benchmarks (PMLB)~\cite{romano2021pmlb}, while the latter are $130$ datasets generated with physics equations from the Feynman Lectures~\cite{feynman2006feynman, feynman2015feynman} compiled by~\cite{doi:10.1126/sciadv.aay2631}, along with $14$ nonlinear dynamical systems from Strogatz~\cite{strogatz2018nonlinear}. The objective in the ground-truth track is to recover the original equation from noisy observations.

SRBench performs $10$ trials per dataset using a $75/25$ train-test split. For the black-box track, the training is limited at $500,000$ function evaluations or $48$ hours. For the ground-truth track, the limits are $1,000,000$ evaluations or $8$ hours, with varying levels of noise $[0, 0.1, 0.01, 0.001]$ added to the target $\mathbf{y}$.
In our experiments, Brush used fixed hyperparameters: population size of $1000$, $100$ generations, maximum tree depth of $10$, maximum expression size of $128$, $10$ iterations of local optimization, and function set: $+, -, \times, \div, \sin, \cos, \tanh, \exp, \log, \sqrt{\cdot}, \text{pow}, \text{Split}$.

Results for other algorithms were taken from the original SRBench benchmark~\cite{La2025cavalab_srbench}. Additionally, we ran recent relevant SR methods such as E2E~\cite{kamienny_end--end_nodate}, uDSR~\cite{UnifiedFrameworkForDeepSR}, and TPSR+E2E~\cite{shojaee_transformer-based_nodate} using their default configurations.
Although the inclusion of these methods provide only a rough estimate of their performance, we did it out of completeness given their reported performances, as their raw results are not available, potentially limiting more fair comparisons.

\begin{figure}[htb]
    \centering
    \subfloat[Pareto plot comparing model size rank ($y$ axis) and $R^2$ score rank ($x$ axis) for black-box problems. Smaller ranks are better. Points denotes the median, and bars denotes the $95\%$ CI. 
    \label{fig:results-black-box-pareto-plot-rank}]{%
       \includegraphics[width=0.44\linewidth,valign=t]{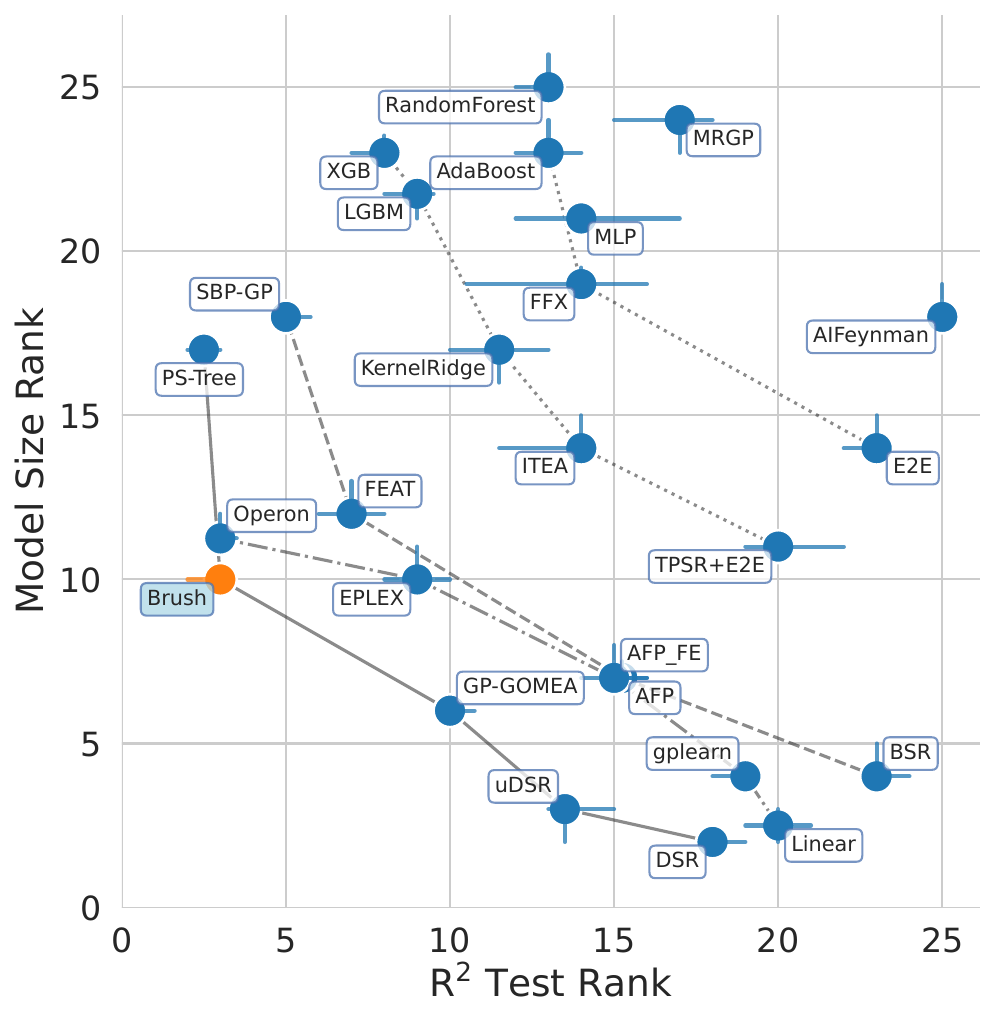}}
    \hfill
    \subfloat[Mean accuracy solution rate for ground-truth problems with varying noise levels, where the rate is defined as the percentage of results with $R^2>0.999$ (four significant figures). Color/shapes indicates noise levels (normal distribution) added to $\mathbf{y}$, and bars denotes the $95\%$ CI.
    \label{fig:results-black-box-pareto-plot-raw}]{%
    \includegraphics[width=0.54\linewidth,valign=t]{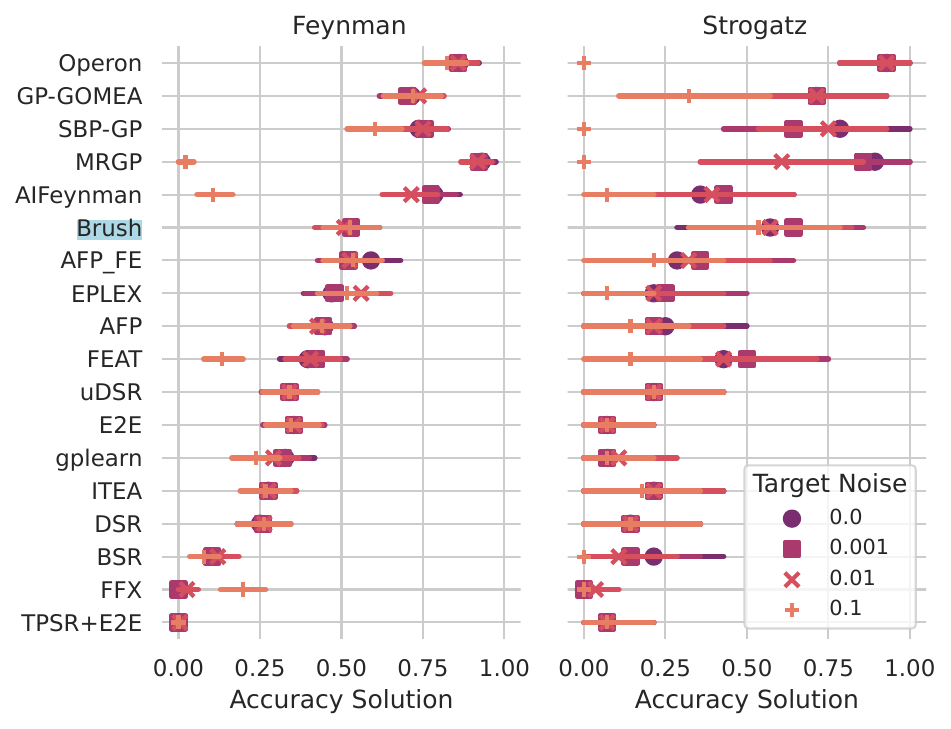}} \\
    \caption{Brush's SRBench results. Bars denotes the bootstrapped confidence intervals (CI).}
    \label{fig:srbench-results}
    \vspace{-2.0em}
\end{figure}

Brush's performance on SRBench is summarized in Figure~\ref{fig:srbench-results}.
The Pareto plot for the black-box problems shows that Brush produces significantly smaller models than PSTree, and often smaller than Operon. It dominates FEAT in both model simplicity and accuracy, achieving the second-best rank in test $R^2$ score. 

For the ground-truth track, Brush recovers accurate expressions across varying noise levels. Although split nodes and weighted terminals may prevent exact equation recovery, we observe that over $50\%$ of solutions achieve $R^2 > 0.999$ with four significant figures for all levels of target noise, and Brush achieves mean (SD) of $84.11 (\pm25.77)\%$ accuracy solution with no noise, and $83.63(\pm25.55)\%$ at maximum noise with three significant figures (\textit{i.e.} $R^2 > 0.99$). Brush is robust to noise, unlike methods such as AIFeynman and MRGP, which show a significant drop in accuracy as noise increases. This suggests Brush successfully recovers approximately half of all Feynman and Strogatz equations, regardless of noise level.
PS-Tree does not appear in the ground-truth track, as its enforced use of split-wise operations prevents it from modeling exact equations, and their original authors opted out for this track.

These results shows that Brush achieves SotA performance across a diverse set of over $250$ datasets.
Brush matches the accuracy of traditional nonlinear optimization methods, while allowing for flexible expression structures with decision splits at any node.
Overall, Brush demonstrates good potential in regression and physic equation modeling tasks, with robust performance in noisy settings.

\section{Methods}

To demonstrate the split nodes utility, we designed a high risk classification experiment based on clinical risk scores derived from the MIMIC-IV-ED v2.2 dataset.
MIMIC is a publicly available dataset containing electronic medical records from $2008$ to $2019$, collected at Beth Israel Deaconess Medical Center (BIDMC)~\cite{johnson2023mimic}. These data are deidentified and made available to researchers under a data use agreement following completion of human subjects training.
Using the MIMIC-IV-ED pipeline~\cite{xie2022benchmarking}, we extracted one simple clinical calculation (Mean Arterial Pressure (MAP)), the Cardiac Arrest Risk Triage (CART) score, and a simplified version of the Modified Early Warning Score (MEWS). While these scores were originally designed for ordinal assessment, we transformed them into binary classification tasks by using thresholds~\cite{tan2022modified} for high risk of catastrophic deterioration classification.

For MAP we used the standard formula $\text{MAP} = \frac{1}{3}\text{SBP} + \frac{2}{3}\text{DBP}$, where SBP and DBP are the systolic and diastolic blood pressure, respectively.

The CART score (Table~\ref{tab:CART}) aggregates points from respiratory rate, heart rate, DBP, and age, with a total score ranging from $0$ to $57$
We considered the threshold of $\geq 12$ for high risk.

\begin{table}[htb]
    \vspace{-1.0em}
    \caption{CART scoring system \cite{churpek2012derivation}.}
    \label{tab:CART}
    \centering
    \scriptsize
    \addtolength{\tabcolsep}{-0.4em}
    \begin{tabular}{@{}rccccccccc@{}}
    \toprule
    Score & 0 & 4 & 6 & 8 & 9 & 12 & 13 & 15 & 22 \\ \midrule
    Respiratory Rate & $<21$ &  &  & $[21-24)$ &  & $[24-26)$ &  & $[26-29)$ & $\geq29$ \\
    Heart Rate & $<110$ & $[110-140)$ &  &  &  &  & $\geq140$ &  &  \\
    Diastolic BP (mmHg) & $\geq49$ & $[40-50)$ & $[35-40)$ &  &  &  & $\leq35$ &  &  \\
    Age & $<55$ & $[55-70)$ &  &  & $\geq70$ &  &  &  &  \\ \bottomrule
    \end{tabular}
\end{table}

The simplified MEWS score (Table~\ref{tab:MEWS}) combines SBP, heart rate, respiratory rate, temperature, and a responsiveness score (AVPU). 
We excluded the AVPU component, because it was not accessible in a structured form. We used an interpretive threshold of $\geq 3$. The maximum attainable score in our setting is $11$.

\begin{table}[htb]
    \vspace{-1.0em}
    \caption{Simplified MEWS scoring system \cite{subbe2001validation}.}
    \label{tab:MEWS}
    \centering
    \scriptsize
    \addtolength{\tabcolsep}{-0.4em}
    \begin{tabular}{@{}rccccccc@{}}
    \toprule
     & 3 & 2 & 1 & 0 & 1 & 2 & 3 \\ \midrule
    Systolic BP (mmHg) & $<71$ & $[71-81)$ & $[81-101)$ & $[101-200)$ &  & $\geq200$ &  \\
    Heart Rate &  & $<41$ & $[41-51)$ & $[51-101)$ & $[101-111)$ & $[111-130)$ & $\geq130$ \\
    Respiratory Rate &  & $<9$ &  & $[9-15)$ & $[15-21)$ & $[21-30)$ & $\geq30$ \\
    Temperature ($^\circ$C) &  & $<35$ &  & $[35-38.5)$ &  & $\geq38.5$ &  \\ \bottomrule
    \end{tabular}
\end{table}

For each score, we trained models to solve both the original regression task and the binary classification task based on the clinical thresholds. We compared Brush with interpretable machine learning (ML) baselines and symbolic regression methods including split-wise operations: PS-Tree, FEAT, decision trees (DT), and logistic regression with L2 regularization (LR L2). We focused comparisons on PS-Tree due to its structural similarity to Brush (both rely on split-wise expressions), and on FEAT due to its successful application in clinical decision-making settings.

Hyperparameter settings were as follows.
DT had no limit on maximum depth. LR used L2 regularization and the \texttt{liblinear} solver. Random forests were fine-tuned for each run to optimize the number of estimator by searching the parameter grid: \texttt{n\_estimators} $\in \{10, 100, 1000\}$, \texttt{min\_weight\_fraction\_leaf} $\in \{0.0, 0.25, 0.5\}$, and \texttt{max\_features} $\in \{\text{``sqrt''}, \text{``log2''}, \text{None}\}$.
FEAT was tuned with population sizes $\in \{100, 500, 1000\}$, generation steps $\in \{250, 500, 2500\}$, and learning rates $\in \{0.1, 0.3\}$.
Preliminary experiments with Brush used default population initialization which includes a wide array of operators; these experiments were used to generate example models for clinical interpretation. 
Subsequent experiments, for which we report aggregate performance metrics, initialized the population with decision tree split nodes, and used the same other following fixed hyperparameters: \texttt{pop\_size} = 500, \texttt{max\_gens} = 100, \texttt{max\_depth} = 12, \texttt{max\_size} = 100, and function set \{\texttt{Add}, \texttt{Sub}, \texttt{Mul}, \texttt{Div}, \texttt{Ceil}, \texttt{Floor}, \texttt{Pow}, \texttt{Log}, \texttt{Min}, \texttt{Max}, \texttt{Split}\}.

We conducted $5$ runs of $5$-fold cross-validation, totaling $25$ runs per method. Each dataset included $10,000$ samples split in a stratified $75/25$ train-test split. 
Class imbalance was present across tasks, with positive cases prevalence of $0.09$ for CART, and $0.11$ for MEWS. Due to this imbalance, we report the Area Under the Precision-Recall Curve (AUPRC) using the average precision score calculation, a more suitable measure for inbalanced class problems.
The feature set comprised $77$ variables including: triage vital signs (temperature, heart rate, blood pressure), emergency department (ED) visits in the past $30$ or $365$ days, chief complaints, and Charlson Comorbidity Index (CCI). We excluded post-admission ED vitals, which could act as proxies for triage admission.
Even that the triage scores relies on at most $5$ features, we kept the entire set to challenge the algorithm's feature selection when building interpretable models.

\section{Results}

We evaluated the performance of all methods on both the scoring and classification problems using the triage scores.
In the scoring version, the objective was to predict the clinical score values (MAP, CART, simplified MEWS), formulated as a regression problem. Table~\ref{tab:scoring_metrics} presents the average $R^2$ values and model size for each method, along with standard deviations (SD) across the runs.
Model size is defined as the number of nodes required to represent the model in a Brush-style expression tree.

\begin{table}[htb]
    \caption{
    Average $R^2$ and model size for each scoring task ($\pm$ SD).
    }
    \label{tab:scoring_metrics}
    \centering
    \scriptsize
    \addtolength{\tabcolsep}{-0.4em}
    \begin{tabular}{@{}rrrrrrr@{}}
    \toprule
     & \multicolumn{2}{c}{MAP} & \multicolumn{2}{c}{CART Score} & \multicolumn{2}{c}{Simplified MEWS Score} \\\cmidrule(l){2-3} \cmidrule(l){4-5}\cmidrule(l){6-7}  
     & $R^2$ & Size & $R^2$ & Size & $R^2$ & Size \\ \midrule
    RF & $0.84 \pm 0.28$ & $(3.50 \pm 4.11) \times 10^6$  & $0.99 \pm 0.00$ & $(1.58 \pm 0.55) \times 10^6$ & $0.98 \pm 0.01$ & $(1.19 \pm 0.77) \times 10^5$ \\
    DT & $0.81 \pm 0.26$ & $11662.24 \pm 89.93$ &$ 0.99 \pm 0.00$ & $194.68 \pm 6.93$ & $0.98 \pm 0.01$ &$ 206.44 \pm 8.40$ \\
    LR L2 & $1.00 \pm 0.00$ & $303.00 \pm 0.00$ & $0.56 \pm 0.11$ & $303.00 \pm 0.00$ & $0.43 \pm 0.08$ & $303.0 \pm 0.00$ \\ \midrule
    PSTree & $1.00 \pm 0.00$ & $217.64 \pm 80.13$ & $0.87 \pm 0.18$ & $813.12 \pm 141.54$ & $0.85 \pm 0.12$ & $815.36 \pm 169.99$ \\
    FEAT & $1.00 \pm 0.00$ & $13.48 \pm 4.99$ & $0.71 \pm 0.23$ & $47.76 \pm 21.42$ & $0.65 \pm 0.10$ & $31.28 \pm 20.13$ \\
    Brush & $1.00 \pm 0.00$ & $13.16 \pm 4.34$ & $0.82 \pm 0.26$ & $76.44 \pm 15.17$ & $0.74 \pm 0.04$ & $45.00 \pm 20.42$ \\
    \bottomrule
    \end{tabular}
\end{table}

Next, we report the classification versions of the CART and simplified MEWS tasks, where models predict whether a patient is at high risk of deterioration based on different triage scores. 
Table~\ref{tab:heuristic_metrics} shows the average AUPRC and model size.
Due to its lack of native support for classification, PS-Tree is excluded from this comparison.

\begin{table}[htb]
    \caption{Average AUPRC and size ($\pm$ SD) for each of the clinical decision problems.
    }
    \label{tab:heuristic_metrics}
    \centering
    \scriptsize
    \addtolength{\tabcolsep}{-0.4em}
    \begin{tabular}{@{}rrrrr@{}}
    \toprule
     & \multicolumn{2}{c}{CART deterioration} & \multicolumn{2}{c}{Simplified MEWS deterioration} \\\cmidrule(l){2-3} \cmidrule(l){4-5}
     & AUPRC & Size & AUPRC & Size \\ \midrule
    RF & $1.00 \pm 0.01$ & $(8.39 \pm 15.01) \times 10^3$ & $0.99 \pm 0.00$ & $(2.22 \pm 2.56) \times 10^4$ \\
    DT & $0.99 \pm 0.01$ & $38.48 \pm 3.11$ & $0.99 \pm 0.01$ & $63.76 \pm 2.86$ \\
    LR L2 & $0.75 \pm 0.02$ & $303.00 \pm 0.00$ & $0.81 \pm 0.03$ & $303.00 \pm 0.00$ \\ \midrule
    FEAT & $0.69 \pm 0.17$ & $60.76 \pm 77.74$ & $0.89 \pm 0.03$ & $55.60 \pm 27.82$ \\
    Brush & $0.99 \pm 0.01$ & $73.80 \pm 15.18$ & $0.95 \pm 0.02$ & $61.84 \pm 11.63$ \\
    \bottomrule
    \end{tabular}
\end{table}

We executed the experiments using Brush without split nodes to assess its impact on model performances. Disabling split nodes did not show statistically significant impact on both size and $R^2$ scores for the regression problems, but showed significant degradation to AUPRC for the deterioration classification tasks. On the test set, Brush without split nodes achieves a smaller AUPRC compared to its counterpart with split nodes in both CART Deterioration ---$0.78\pm 0.07$ \textit{versus} $0.99\pm 0.01$ (p$\leq$1.00e-04)--- and MEWS Deterioration tasks --- $0.88\pm 0.04$ \textit{versus} $0.95\pm 0.02$ (p $\leq$ 1.00e-04). Using the split nodes didn’t hurt the performance for other problems. This implies the gains are not only in interpretability, but also in classification performance.

\begin{figure}[htb]
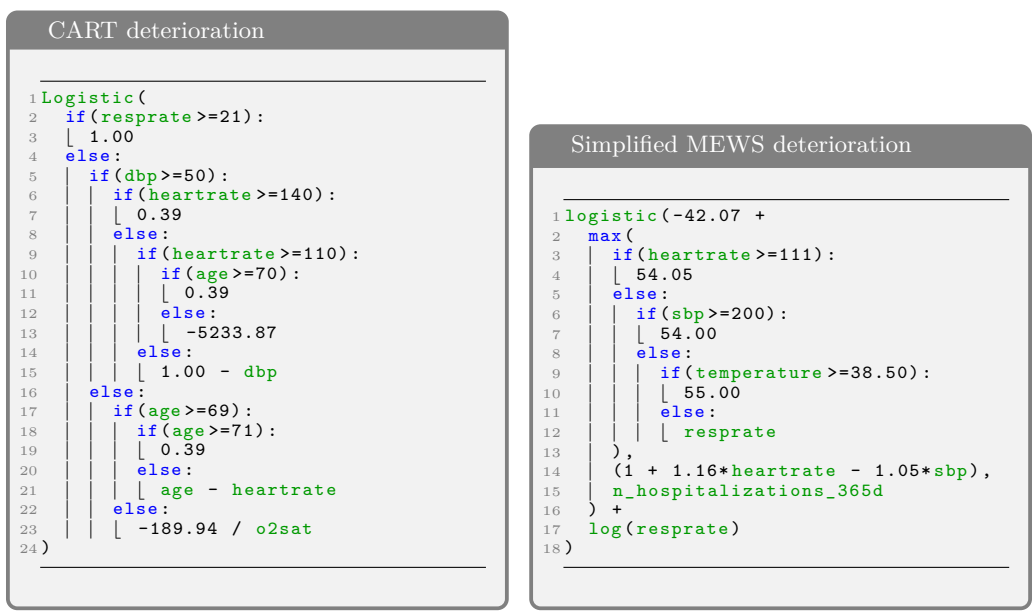

    \centering
    \begin{minipage}[t]{0.49\textwidth}
        \centering
        \begin{tcolorbox}[colframe=black!50, colback=gray!10, title=CART deterioration]
            \lstset{language=Python, 
                basicstyle=\linespread{0.9}\ttfamily\scriptsize, 
                numberstyle=\tiny\color{gray},
                keywordstyle=\color{blue},
                commentstyle=\color{dkgreen},
                stringstyle=\color{mauve},
                showstringspaces=false,
                numbersep=1pt,
                xleftmargin=-0.1cm,
                numbers=left,
                xrightmargin=-0.25cm,
                frame=bt,
                identifierstyle=\color{dkgreen}}
\begin{lstlisting}[mathescape=true]
Logistic(
  if(resprate>=21):
  $\lfloor$  1.00
  else:
  $\mid$  if(dbp>=50):
  $\mid$  $\mid$  if(heartrate>=140):
  $\mid$  $\mid$  $\lfloor$  0.39
  $\mid$  $\mid$  else:
  $\mid$  $\mid$  $\mid$  if(heartrate>=110):
  $\mid$  $\mid$  $\mid$  $\mid$  if(age>=70):
  $\mid$  $\mid$  $\mid$  $\mid$  $\lfloor$  0.39
  $\mid$  $\mid$  $\mid$  $\mid$  else:
  $\mid$  $\mid$  $\mid$  $\mid$  $\lfloor$  -5233.87
  $\mid$  $\mid$  $\mid$  else:
  $\mid$  $\mid$  $\mid$  $\lfloor$  1.00 - dbp
  $\mid$  else:
  $\mid$  $\mid$  if(age>=69):
  $\mid$  $\mid$  $\mid$  if(age>=71):
  $\mid$  $\mid$  $\mid$  $\lfloor$  0.39
  $\mid$  $\mid$  $\mid$  else:
  $\mid$  $\mid$  $\mid$  $\lfloor$  age - heartrate
  $\mid$  $\mid$  else:
  $\mid$  $\mid$  $\lfloor$  -189.94 / o2sat
)
\end{lstlisting}
        \end{tcolorbox}
    \end{minipage}
    \hfill
    \begin{minipage}[b]{0.49\textwidth}
        \centering
        \begin{tcolorbox}[colframe=black!50, colback=gray!10, title=Simplified MEWS deterioration]
            \lstset{language=Python, 
                basicstyle=\linespread{0.9}\ttfamily\scriptsize, 
                numberstyle=\tiny\color{gray},
                keywordstyle=\color{blue},
                commentstyle=\color{dkgreen},
                stringstyle=\color{mauve},
                numbers=left,
                showstringspaces=false,
                numbers=left,
                numbersep=1pt,
                xleftmargin=-0.1cm,
                xrightmargin=-0.25cm,
                frame=bt,
                identifierstyle=\color{dkgreen}}
\begin{lstlisting}[mathescape=true]
logistic(-42.07 + 
  max(
  $\mid$  if(heartrate>=111):
  $\mid$  $\lfloor$  54.05
  $\mid$  else:
  $\mid$  $\mid$  if(sbp>=200):
  $\mid$  $\mid$  $\lfloor$  54.00
  $\mid$  $\mid$  else:
  $\mid$  $\mid$  $\mid$  if(temperature>=38.50):
  $\mid$  $\mid$  $\mid$  $\lfloor$  55.00
  $\mid$  $\mid$  $\mid$  else:
  $\mid$  $\mid$  $\mid$  $\lfloor$  resprate
  $\mid$  ),
  $\mid$  (1 + 1.16*heartrate - 1.05*sbp),
  $\mid$  n_hospitalizations_365d
  ) + 
  log(resprate)
)
\end{lstlisting}
        \end{tcolorbox}
    \end{minipage}
    \caption{Exemplar Brush models edited as Python code for catastrophic deterioration prediction using two different scoring systems CART and simplified MEWS.}
    \label{fig:final-brush-models}
    \vspace{-1.0em}
\end{figure}

To further explore the interpretability of Brush models for clinical decision-making, we selected exemplar models from the classification tasks. 
We note that this model interpretation was conducted using the preliminary Brush settings that did not include split node seeding of the initial population. 
First, we filtered out models with an AUPRC below $0.90$. 
We also observed that some models with high AUPRC exhibited a balanced accuracy of $0.5$ on the test partition, indicating random performance. 
Consequently, we excluded models with a balanced accuracy below $0.9$. From the remaining models, we selected the smallest, which is visualized in Figure \ref{fig:final-brush-models}.

\begin{figure}[htb]
    \centering
    \subfloat[$R^2$ score and model size for predicting the original scoring values. \label{fig:results-scoring-barplot}]{%
       \includegraphics[width=0.59\linewidth,valign=t]{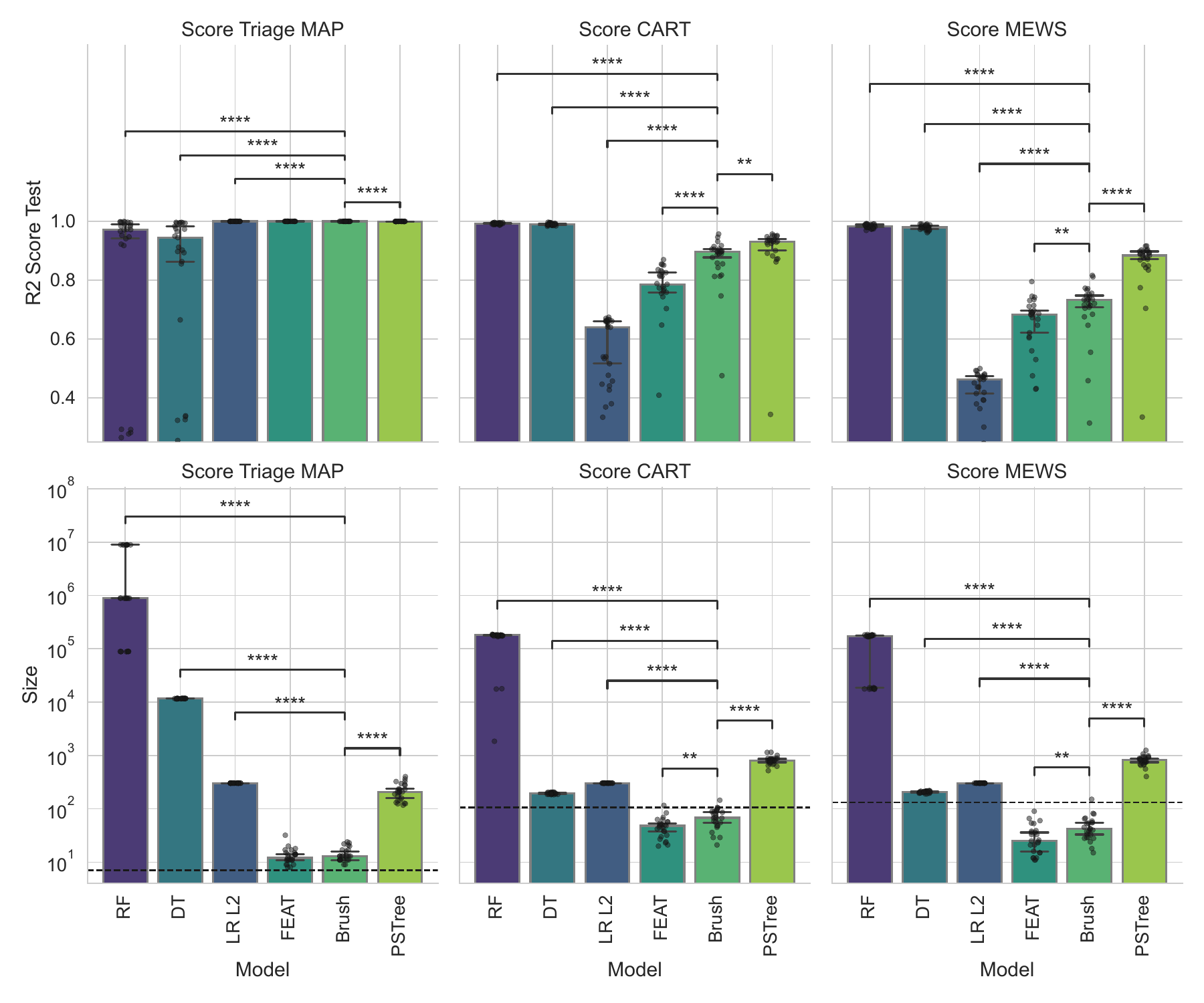}}
    \hfill
    \subfloat[AUPRC and size for predicting deterioration. \label{fig:results-deterioration-barplot}]{%
    \includegraphics[width=0.395\linewidth,valign=t]{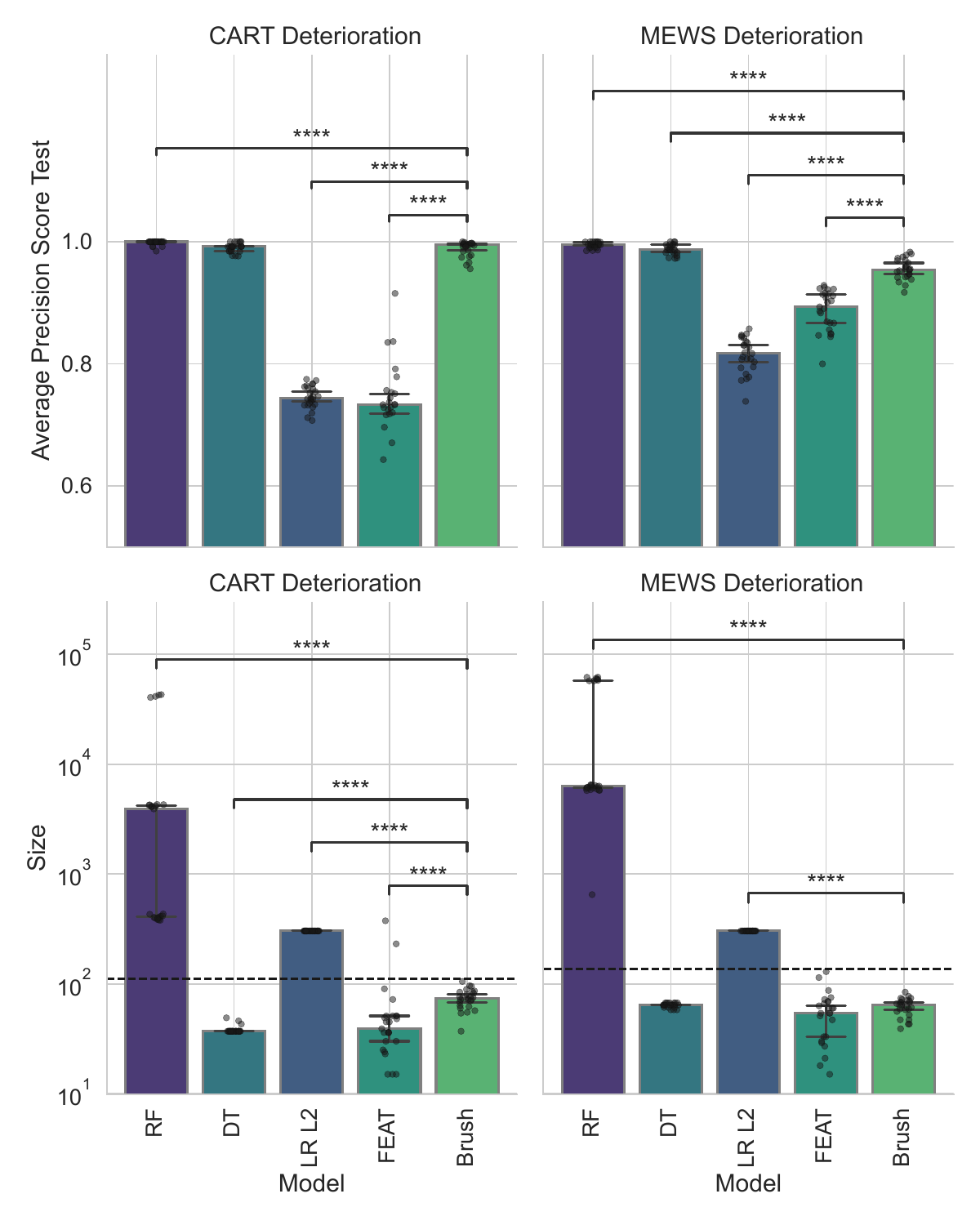}} \\
    \caption{Metrics and sizes for the clinical decision experiment. The size of the original decision model is denoted as a dashed line.
    Statistical comparisons are conducted using Mann-Whitney-Wilcoxon two-sided test with Holm-Bonferroni corrections. 
       *: 1.00e-02 $<$ p $\leq$ 5.00e-02; **: 1.00e-03 $<$ p $\leq$ 1.00e-02 ***: 1.00e-04 $<$ p $\leq$ 1.00e-03 ****: p $\leq$ 1.00e-04.
    Non-significant comparisons were omitted.
    }
    \label{fig:barplot_heuristics}
    \vspace{-1.0em}
\end{figure}

To complement the tabular results in Table~\ref{tab:heuristic_metrics}, Figure~\ref{fig:barplot_heuristics} presents barplots of both performance and model size for the classification task.
Subfigure~\ref{fig:results-scoring-barplot} displays the $R^2$ scores and complexities for the regression-based scoring tasks, and Subfigure~\ref{fig:results-deterioration-barplot} shows the AUPRC scores and complexities for the classification tasks. These plots also include statistical comparisons between Brush and the other methods using the Mann-Whitney-Wilcoxon two-sided test with Holm-Bonferroni correction. 
We set $y$ axis lower bound to $0.5$ in AUPRC plots, and $0.25$ in $R^2$ plots.

\section{Discussion}~\label{sec:discussion}

We observe that the MAP score —a weighted sum of two features— showed to be a particular challenge for traditional ML methods. Tree-based models such as RF and DT performed poorly compared to other methods, often producing large models with greater variability. This is likely due to the continuous nature of the output, making it difficult for split-based methods to perform well due to their inherent discretization mechanism with constants as leaves. LR L2 also found large and less interpretable models, failing on the feature selection aspect. PSTree consistently produced overly large models, possibly due to the high dimensionality of the feature space combined with its partitioning strategy, which creates numerous splits. Alternatively, PSTree result may be due its reliance on random guessing the coefficients. In contrast, FEAT and Brush produced more compact models with similar or better performance.

For the CART and simplified MEWS scoring tasks, we notice that LR L2 shows a drop in $R^2$, while FEAT and Brush achieved superior performance. This suggests that logistic regression, when combined with the feature selection and feature engineering capabilities provided by genetic programming (as used in both symbolic regression methods), can outperform classical logistic regression. For these specific tasks, Brush consistently outperforms FEAT, at the cost of slightly larger models. PSTree achieves higher performance than Brush, but at the cost of models that are more than $20$ times larger.

Figure \ref{fig:results-scoring-barplot} provides statistical comparisons, showing that Brush models showed statistically significant differences except when compared to FEAT in the MAP score. It also reveals that the trade-off between AUPRC and model size is overall unfavorable for PSTree due to its extremely large model sizes.
Brush models showed relatively low variance across runs when compared to other SR methods, indicating stable behavior.

When model predictions are transformed into clinical decision tasks, we observe notable performance differences in Table~\ref{tab:heuristic_metrics}. In particular, both SR methods perform worse than DT on rule-based classification tasks. Compared to FEAT, Brush demonstrates superior performance on CART but yields comparable results on simplified MEWS. These differences are also present in regression tasks; however, Brush and FEAT generate more compact models than shallow methods such as RF, DT, and LR L2.
Brush outperforms LR L2 in all cases, and produces smaller models than DT for simplified MEWS. Statistical comparisons in Figure~\ref{fig:results-deterioration-barplot} show no significant size difference between Brush and DT for CART, but a significant reduction for MEWS.

Overall, we position Brush as a strong compromise between the three main modeling approaches. It approximates DT performance in classification tasks while yielding better regression models; it enhances logistic modeling compared to LR L2 through feature engineering and selection using genetic programming; and it outperforms FEAT and PSTree in different aspects by combining split-wise operations with parameter optimization. Allowing specific customization in the trade-off between performance and complexity could yield better task-specific performance.
The Brush model uses a logistic root, meaning it outputs a probability estimate for deterioration risk. This probabilistic output can be evaluated using metrics like area under the Precision-Recall Curve or ROC curve, and an optimal classification threshold can be chosen accordingly.

For the CART deterioration task, the model depicted in Figure~\ref{fig:final-brush-models} has a size of $62$ nodes, an AUPRC of $0.92$, and a balanced accuracy of $0.99$ on the test set.
The model evaluation begins with a split on respiratory rate ($>21$), and can be seen as the most immediate high-risk indicator, identifying high risk immediately --- consistent with the original scoring system. If this condition fails, the model proceeds through a nested evaluation of diastolic blood pressure (DBP), heart rate, and age, computing a final probability of deterioration. 
We can see combinations of DBP and heart rate in the first block of conditionals, which are the two factors other than respiration rate that have the highest scores in the CART scoring system of DBP $\leq35$ and heart rate $\geq140$, resulting in a risk higher than the threshold, thus a positive label. Age has a smaller priority for probability modelling, and, in case the patient is not at advanced ages above $70$, then a computation is made using `age - heart rate'. We note that the final branch --- when all conditionals fail, meaning a superior bound was set to all tested features ---, references $O_2$ saturation, which was not used during label generation and may be an artifact, and could be simplified. This structure mirrors the kind of heuristic reasoning a clinician might follow and resembles the CART scoring system, although in a compact, rule-based form.

The Brush model for simplified MEWS deterioration prediction had an AUPRC of $0.96$ and a balanced accuracy of $0.91$ on the test set, with a model size of $68$ nodes.
Impressively, it identified the relevant features aligned with the actual simplified MEWS scoring system from a pool of $77$ possible features, and learned threshold values consistent with the actual clinical model. The model uses a \texttt{Max} function to return a fixed output associated with high-risk prediction if any of the vital signs (heart rate, SBP, or temperature) are critically high. If none of these are critical, the model considers a linear combination of heart rate and SBP, which identifies a combination of relative tachycardia and hypotension, and number of previous admissions. Finally this is added to the logarithm of respiratory rate. This algorithm integrates risk across vital signs in several ways, such that it is simpler than the MEWS score which treats each element as independent.

\section{Conclusions}~\label{sec:conclusions}

In this paper we propose Brush, a Multi-Objective symbolic regression algorithm specially designed for solving problems where split-wise functions are desired.
The novelty introduced by Brush is integrating split-wise functions with non-linear optimization methods, also combining several state-of-the-art components into a genetic programming framework, namely $\epsilon$-lexicase selection, NSGA-II survival, weighted nodes, fast simplification methods, and classification support. Brush contributes to building interpretable and high-performing models.

We shown that Brush achieves Pareto-optimal performance on a collection of $122$ real world datasets, and successfully retrieves more than half of the Feynman and Strogatz problems, even in noisy scenarios. These results highlight its potential as a tool for deriving interpretable and high-performing heuristics in healthcare and other domains where interpretability is essential.

Applied to real-world electronic health records data, Brush consistently identified relevant features and produced compact decision models with near-perfect predictive performance. 
Overall, the models are simple, interpretable, and captures the essential structure of clinical scoring with minimal deviation, missing only some corner cases. Brush effectively balances interpretability, performance, and stability ---rarely achieved simultaneously in clinical modeling tools.

We notice some limitations. First, Brush was not fine tuned, which could lead to better results, but was out of the scope of this paper. Second, Brush minimizes linear complexity based on arbitrarily defined complexities for each node, which could be tweaked or derived from prior distributions observed in different fields. 

\subsubsection*{Data availability}


SRBench \cite{srbench} experiments and results are available at GitHub  \cite{La2025cavalab_srbench}.
MIMIC-IV-ED dataset \cite{mimic-ed-doi} is available at Physionet \url{https://physionet.org/content/mimic-iv-ed/2.2/}.
MIMIC-IV-ED pipeline \cite{gupta2022extensive} is available at GitHub \cite{mehak252024healthylaife}.
Brush source code \cite{brush2025cavalab} is hosted at GitHub \url{https://github.com/cavalab/brush/}.
Our experiments are available at GitHub \url{https://github.com/cavalab/brush_paper_experiments}.

\subsubsection*{Acknowledgments}
W.G.L., D.S.H, and G.S.I.A.~are supported by Patient Centered Outcomes Research Institute (PCORI) ME-2020C1D-19393.
The statements in this work are solely the responsibility of the authors and do not necessarily represent the views of the Patient-Centered Outcomes Research Institute (PCORI), its Board of Governors or Methodology Committee.
F.O.F.~is supported by Conselho Nacional de Desenvolvimento Cient\'{i}fico e Tecnol\'{o}gico (CNPq) grant 301596/2022-0.
G.S.I.A.~is supported by Coordena\c{c}\~{a}o de Aperfei\c{c}oamento de Pessoal de N\'{i}vel Superior (CAPES) finance Code 001 and grant 88887.802848/2023-00.
J.D.R.~is supported by US National Libraries of Medicine grant R00-LM013646.




{\small
\bibliographystyle{RS}
\bibliography{refs}
}

\end{document}